%% file: main.tex
\documentclass[sigconf]{aamas}  
\usepackage{balance}  

\usepackage{booktabs}

\setcopyright{ifaamas}  
\acmDOI{}  
\acmISBN{}  
\acmConference[AAMAS'19]{Proc.\@ of the 18th International Conference on Autonomous Agents and Multiagent Systems (AAMAS 2019)}{May 13--17, 2019}{Montreal, Canada}{N.~Agmon, M.~E.~Taylor, E.~Elkind, M.~Veloso (eds.)}  
\acmYear{2019}  
\copyrightyear{2019}  
\acmPrice{}  

\usepackage{csquotes} 
\usepackage{relsize} 
\usepackage{amssymb} 
\usepackage[nice]{nicefrac}
\usepackage[free-standing-units]{siunitx} 
\usepackage{soul} 
\usepackage{colortbl} 
\usepackage{graphicx,color}
\usepackage{enumerate}
\usepackage{booktabs} 
\usepackage{enumitem} 
\usepackage{numprint}
\npthousandsep{,}\npthousandthpartsep{}\npdecimalsign{.}
\usepackage{icomma}
 
\newcommand{\rc}{{RoboCup@Work}}

\begin{document}

\title{Fully Convolutional One--Shot Object Segmentation \\ for Industrial Robotics}   


\author{Benjamin Schnieders\textsuperscript{\dag}}\thanks{This paper is dedicated to the memory of our wonderful colleague and friend Benjamin Schnieders, a bright young scientist, who recently passed away.}
\affiliation{%
  \institution{University of Liverpool}
  \city{Liverpool} 
  \state{United Kingdom}
  \postcode{L69 3BX}
}
\email{Benjamin.Schnieders@liverpool.ac.uk}
\author{Shan Luo}
\affiliation{%
  \institution{University of Liverpool}
  \city{Liverpool} 
  \state{United Kingdom}
  \postcode{L69 3BX}
}
\email{Shan.Luo@liverpool.ac.uk}
\author{Gregory Palmer}
\affiliation{%
  \institution{University of Liverpool}
  \city{Liverpool} 
  \state{United Kingdom}
  \postcode{L69 3BX}
}
\email{G.J.Palmer@liverpool.ac.uk}
\author{Karl Tuyls}
\affiliation{%
  \institution{University of Liverpool}
  \city{Liverpool} 
  \state{United Kingdom}
  \postcode{L69 3BX}
}
\email{K.Tuyls@liverpool.ac.uk}

\begin{abstract}  
The ability to identify and localize new objects robustly and effectively is vital for robotic grasping and manipulation in warehouses or smart factories. Deep convolutional neural networks (DCNNs) have achieved the state-of-the-art performance on established image datasets for object detection and segmentation. However, applying DCNNs in dynamic industrial scenarios, e.g., warehouses and autonomous production, remains a challenging problem. DCNNs quickly become ineffective when tasked with detecting objects that they have not been trained on. Given that re-training using the latest data is time consuming, DCNNs cannot meet the requirement of the \emph{Factory of the Future (FoF)} regarding rapid development and production cycles. To address this problem, we propose a novel one-shot object segmentation framework, using a fully convolutional Siamese network architecture, to detect previously unknown objects based on a single prototype image. We turn to multi-task learning to reduce training time and improve classification accuracy. Furthermore, we introduce a novel approach to automatically cluster the learnt feature space representation in a weakly supervised manner. We test the proposed framework on the RoboCup@Work dataset, simulating requirements for the \emph{FoF}. Results show that the trained network on average identifies $73\percent$ of previously unseen objects correctly from a single example image. Correctly identified objects are estimated to have a $87.53\percent$ successful pick-up rate. Finally, multi-task learning lowers the convergence time by up to $33\percent$, and increases accuracy by $2.99\percent$.

\end{abstract}

%


\maketitle


\input{body}


\bibliographystyle{ACM-Reference-Format}  
\balance  
\bibliography{objdetect2}  

\end{document}

%% file: body.tex
\section{Introduction}
The Factory of the Future~\cite{lasi2014industry} imagines interconnected robots working alongside and in cooperation with humans.
Tasks to be performed range from being in direct contact with humans~\cite{harant2018comparison} during assistance with repetitive or potentially dangerous tasks, to completely autonomous operations, in which mobile robots connect various other, pre-existing robotic workflows. To ensure these autonomous operations run without disturbance for as long as possible, detecting objects and their pickup point robustly is of high priority~\cite{lucke2008smart}.

While Deep Convolutional Neural Networks (DCNNs) form the state-of-the-art in object detection~\cite{schmidhuber2015deep}, they typically suffer from long training periods.
Rapid prototyping production, as in the Factory of the Future, can require turnaround times shorter than the training times of common neural networks. 
In this situation, the robot may be required to pick up produced objects before the network has finished training, halting the entire workflow.
While many techniques exist to speed up the convergence of neural networks~\cite{ren2017faster}, dataset acquisition and labeling poses an additional delay. Thus, for workflows of a smaller batch size, training a robot to detect the produced objects can be infeasible due to temporal constraints.

Training a network to generalize the learned knowledge of recognizing pre-trained objects to correctly recognize untrained classes is the subject of \emph{few-}, \emph{one-}, and \emph{zero-}shot object detection ~\cite{lampert2009learning}. Using \emph{few-} or \emph{one-}shot detection, a robot can be shown one or a few examples of a prototype object, which it can then re-detect during autonomous operations.

\begin{figure}[tb]
	\centering
	\includegraphics[width=\linewidth]{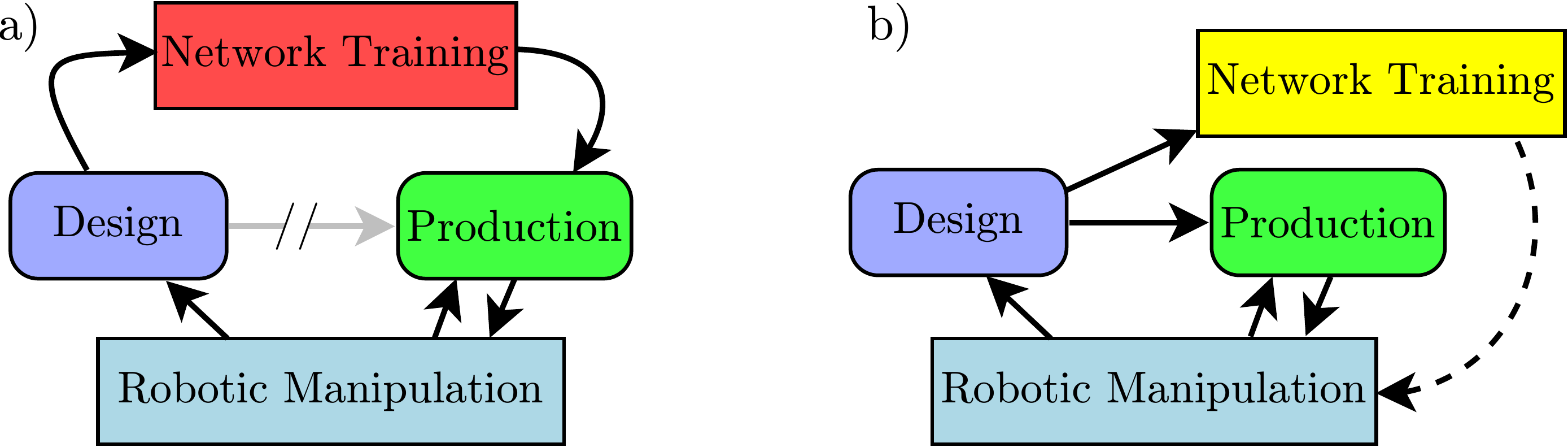}
	\caption{A simplified automated production workflow. a)~Production of newly designed parts has to be delayed until the robot that manipulates them after production is trained. b) The manipulation/production cycle can begin with a one-shot detection network, while the new information is learned. Once training is complete, the robot can use the updated network.}
	\label{fig:workflow}	
\end{figure}

Figure~\ref{fig:workflow} showcases a model workflow of rapid prototyping production. Training the network becomes a bottleneck, as production has to be delayed until the robot can learn to detect, and thus to manipulate, the produced objects. Using a one-shot detection approach alleviates the bottleneck, and thus allows for faster design/production cycles.

While object detection networks are commonly trained to produce bounding boxes around the detected objects~\cite{ren2017faster,redmon2017yolo9000}, object segmentation approaches segment the image pixelwise into semantic classes, i.e., which object class each pixel belongs to. This segmentation output allows the robot to compute an object \emph{pickup point}, and subsequently, to pick it up~\cite{schnieders2018fast}.
The state-of-the-art approaches in semantic segmentation typically build on fully convolutional neural networks~\cite{long2015fully}.
Combining the semantic segmentation approach with one-shot learning leads to a one-shot segmentation network.
Such a network is trained to produce a per-pixel classification of whether each pixel shows part of a previously shown prototype object or not.

Research questions addressed in this paper are the following:
\begin{enumerate}[label=\roman*]
\item To what extent can DCNNs extrapolate their learned object segmentation to unknown data?
\item How much does transforming the optimization process into a multi-task problem benefit the training time?
\end{enumerate}

This paper introduces a method to rapidly train a one-shot object segmentation network to detect industrial objects based on a single prototype image. The method is validated on the new, publicly available\footnote{\url{http://wordpress.csc.liv.ac.uk/smartlab/objectdetectionwork/}} {\rc} dataset~\cite{schnieders2018fast}, which includes objects present in a Factory of the Future simulation~\cite{kraetzschmar2014robocup}.
Results illustrate that our network architecture can identify previously unseen objects in a Factory of the Future simulation with $73\percent$ accuracy after being trained for only \numprint{7000} iterations. As little as \numprint{4000} iterations, or approximately \numprint{57} minutes of training on a NVIDIA 1080 Ti, suffice to re-train the network with all information, producing a $94\percent$ accuracy.

Our main contributions can thus be summarized as follows:
\begin{enumerate}[label=\roman*]
\item A novel framework is proposed for rapidly training Siamese network based, fully convolutional networks to perform one-shot segmentation in industrial scenarios\footnote{We make our code available online: \url{https://github.com/smARTLab-liv/ObjectDetect}};
\item An innovative error metric supporting auto-clustering of objects' representations in feature space is introduced to improve convergence speed;
\item An auxiliary network architecture is showcased to automatically combine the different error metrics in order to reduce hyperparameters.
\end{enumerate}

The remainder of this paper is organized as follows: Section~\ref{sec:bg} provides the background on one-shot object detection and segmentation and discusses related work.
Section~\ref{sec:impl} describes the network architecture used, and how the different tasks are trained simultaneously.
Section~\ref{sec:exp} describes the experimental setup.
Finally, Section~\ref{sec:results} presents and compares the experimental results, and Section~\ref{sec:conc} concludes the paper.

\section{Background and Related Work}
\label{sec:bg}
While general one-shot learning is well studied~\cite{salakhutdinov2012one}, one-shot object detection and segmentation is a very recent field of research~\cite{shaban2017one}. While conventional object detection~\cite{guo2016deep} and segmentation~\cite{long2015fully} neural networks operate only on classes the network has been trained on, the one-shot variants extrapolate the learned information to extract a previously unseen object class, of which only one example is shown. Research fields closest related to one-shot segmentation can be roughly subdivided into two categories, each attempting to solve this challenging task in a specific way.


\textbf{Object co-segmentation} is a weakly or entirely unsupervised approach to learn to extract objects based on visual similarity within a class of objects. Typically, image level classes are presented, i.e., the network is trained with the knowledge that presented images belong to the same class, and can learn to detect similar features~\cite{pinheiro2015weakly,papandreou2015weakly,shen2017weakly}. However, while research in this field is able to detect an object based on the input of another image, co-segmentation networks are not designed to abstract the learned knowledge such that it can be transferred to previously unseen objects. Furthermore, the unsupervised nature of the approach requires large datasets and comparatively long training periods.

\textbf{Siamese networks} have been employed to efficiently perform one-shot object classification~\cite{vinyals2016matching} or detection~\cite{keren2018weakly}. Siamese networks are trained to map image data onto a dense feature vector on a much smaller space, in a way that objects from the same category are mapped onto feature space positions close to each other with respect to a given a distance metric. This distance metric can be automatically learned~\cite{schwartz2018repmet}.
Michaelis \textit{et al}.~\cite{michaelis2018one} have shown Siamese networks to be able to perform semantic segmentation of color-coded symbols in cluttered environments.

The work of Shaban \textit{et al}.~\cite{shaban2017one} is the current state-of-the-art method to perform one-shot object segmentation, outperforming competitive approaches on the Pascal VOC dataset.
However, we identify a number of issues preventing it from being employed effectively in a \emph{Factory of the Future} environment:
First, the authors of ~\cite{shaban2017one} assume that there is no overlap in classes between training and test set. While no class of the test set should appear in the training set, the reverse does not hold: Trained classes should be featured as distractors in the evaluation set as well~\cite{michaelis2018one}.
As segmentation networks tend towards extracting known data, this assumption could prohibit any occurrence of trained objects in workflows exploiting one-shot detection, making it unfit in an industrial context.
Secondly, the approach requires a segmentation mask to be applied to the image including the query object. In an autonomous production environment, a human-generated segmentation mask is unlikely to exist.
The third identified issue is the high complexity of the resulting network. While this produces good results on the real-world Pascal VOC dataset, it is likely to suffer from exceedingly long training periods. This assumption is confirmed by~\cite{shaban2017one} stating that memory constraints only allow for a batch size of \numprint{1}, and training is done with a learning rate of $\num{1e-10}$, a combination producing extremely long training periods.
In contrast, our approach is tested on extracting an unknown object class from a set of at least two unknown classes and up to ten known classes. It is able to extract the object to be queried without the need for a human generated segmentation mask. Most importantly, the network architecture proposed in this paper can be trained quickly, with convergence in as little as \numprint{7000} iterations for one-shot detection, or about \numprint{100} minutes of training on a NVIDIA 1080 Ti. If only recollection of known objects is required, re-training to an accuracy of $94\percent$ can be achieved in \numprint{4000} iterations, or \numprint{57} minutes of training.

\section{Approach}
\label{sec:impl}
Our approach is, on its most basic level, composed of a deep convolutional neural network, which is trained on images featuring objects.
For every training iteration, it is presented with an image showing a single object, and an image featuring multiple objects.
As per convention in search and image retrieval~\cite{chum2009geometric}, we call the single object to be retrieved the \emph{needle} object, and the corresponding image the \emph{needle} image.
The second image, featuring the \emph{needle} object amongst multiple others, is called the \emph{haystack} image.
The performance of the DCNN is evaluated on how well it extracts the \emph{needle} object from the \emph{haystack} image.
The Intersection over Union (IoU) error, measuring to which percentage the desired segmentation and the ground truth differ, is commonly used to evaluate segmentation quality, and is used to train the network.
Additionally, the Euclidean distance between predicted object pickup coordinates and the ground truth is measured during training in order to later derive the probability of a robotic pickup operation succeeding~\cite{schnieders2018fast}.

\begin{figure}[tb]
	\centering
	\includegraphics[width=0.9\linewidth]{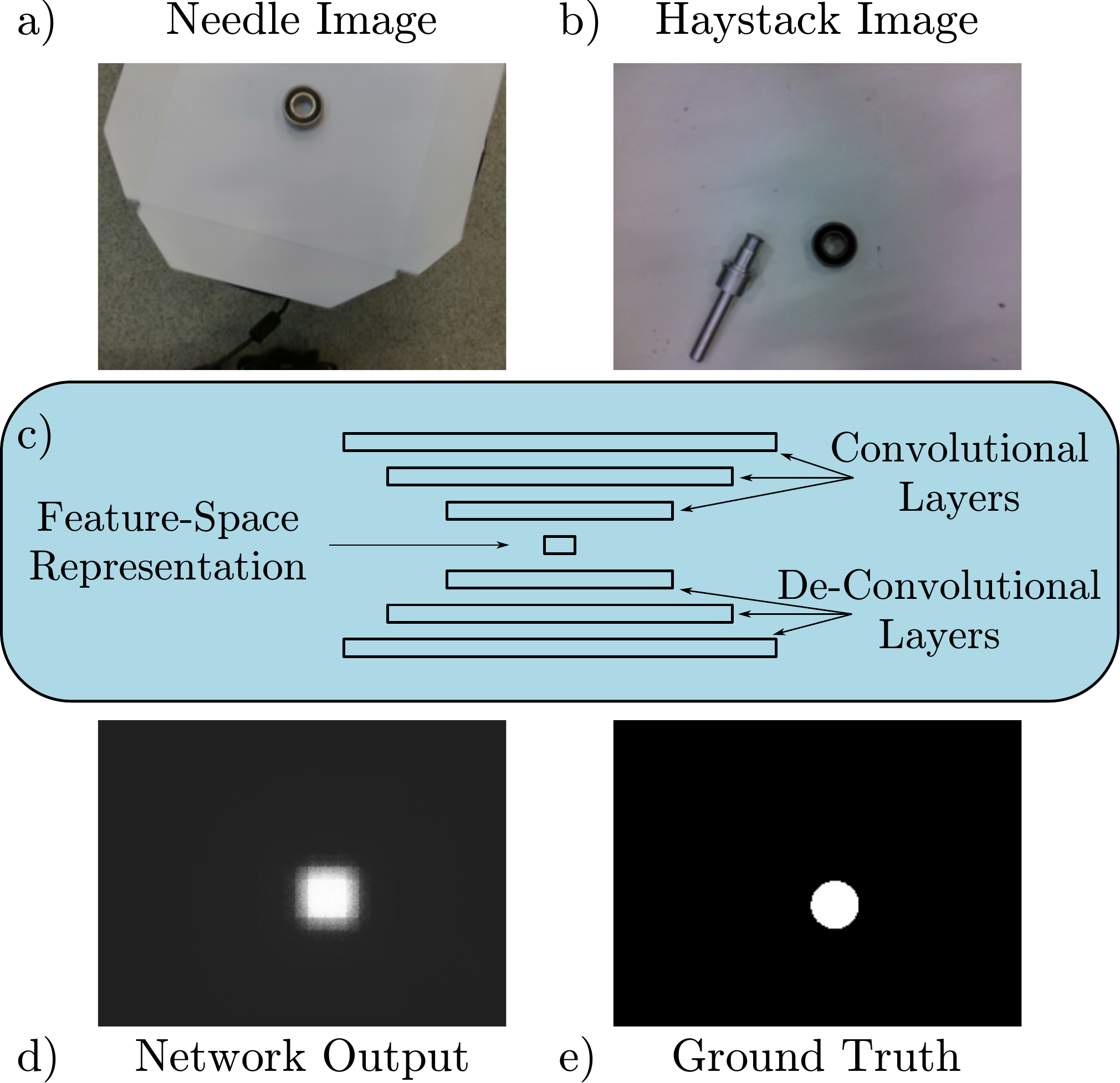}
	\caption{The general principle of a network performing one-shot object segmentation. A \emph{needle} image a) and a \emph{haystack} image b) are presented to the network c), which in turns extracts the \emph{needle} object from the \emph{haystack} image. The network output d) and the ground truth e) are compared and the error is propagated back. Evaluation is performed on object classes the network has not been trained on.}
	\label{fig:schematics}	
\end{figure}

Figure~\ref{fig:schematics} visualizes the nomenclature used on an abstracted segmentation network.
Both \emph{needle} and \emph{haystack} images are presented to the network at the same training iteration. Due to the Siamese nature~\cite{koch2015siamese} of the network used, the feature extraction filters for both images are shared.
Deconvolutional filters restore the original dimensions of the image, and highlight the extracted object. The network output is then compared to the ground truth segmentation, and the error is backpropagated. For the next iteration, a new \emph{needle}/\emph{haystack} combination is selected from the training set. The following section describes the network layout in detail.

\subsection{Network Design}
\label{sec:netDesign}
\begin{figure*}[tb]
	\centering
	\includegraphics[width=0.9\linewidth]{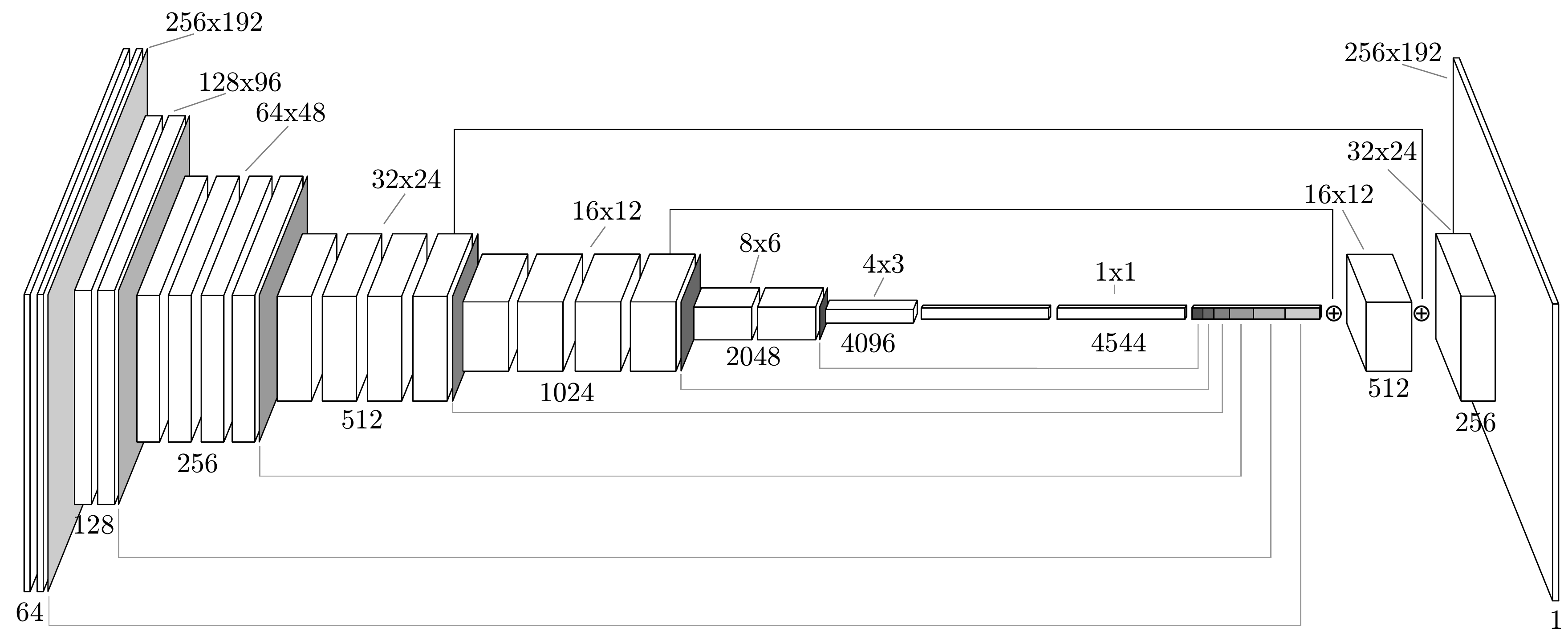}
	\caption{The DCNN used for one-shot object segmentation. Above the network, pixel dimensions are stated; below the network, the number of filters can be found for every stack of convolutional filters. After every such stack, but before the ensuing pooling operation, a \emph{selection} layer is introduced (shaded in grey). Selection layers are inactive when feeding the \emph{needle} image into the network. From the output of the last convolutional 1x1 layer, \emph{selection} layer weights are derived and multiplied onto the result of every filter when the \emph{haystack} image is presented. This can effectively disable all filters potentially extracting unwanted objects, and only keep filters that respond to the \emph{needle} object active.}
	\label{fig:hourglass}	
\end{figure*}

Our one-shot segmentation network operates fully convolutional, without the requirement of fully connected layers or any non-differentiable operations. The DCNN layout is extending~\cite{simonyan2014very} by adding deconvolutional layers as proposed in~\cite{long2015fully}. 
Figure~\ref{fig:hourglass} presents the layout of the DCNN. Convolutional layers successively break down the input image from $256\times192$ pixels and 4 channels to a $1\times1\times4,544$ feature vector representation. The convolutional layers are arranged in eight same-size stacks, after which a $2\times2$ max pooling operation is performed. All filters feature $3\times3$ kernels until a one dimensional feature vector is achieved, after which the eighth stack uses $1\times1$ kernels to introduce additional nonlinearity. All activation functions are leaky ReLUs~\cite{maas2013rectifier}, with $\alpha=0.02$. Dropout layers with a $15\percent$ dropout rate are employed after the third to eighth stack of convolutional layers. Depending on whether the \emph{needle} or \emph{haystack} image is presented, the grey shaded \emph{selector} layers are either ignored, or multiplied onto the outcome of every filter of the layer before.


To perform one-shot object segmentation, we use a Siamese architecture of the feature extraction part of the network, meaning that during one training iteration, two images will be presented to the convolutional layers. However, we do not train a difference metric; instead, we train the network to enable or disable convolutional filters that produce the extraction of the desired object only.
The following paragraph provides a step by step walkthrough of a training iteration.

\begin{figure}[tb]
	\centering
	\includegraphics[width=0.6\linewidth]{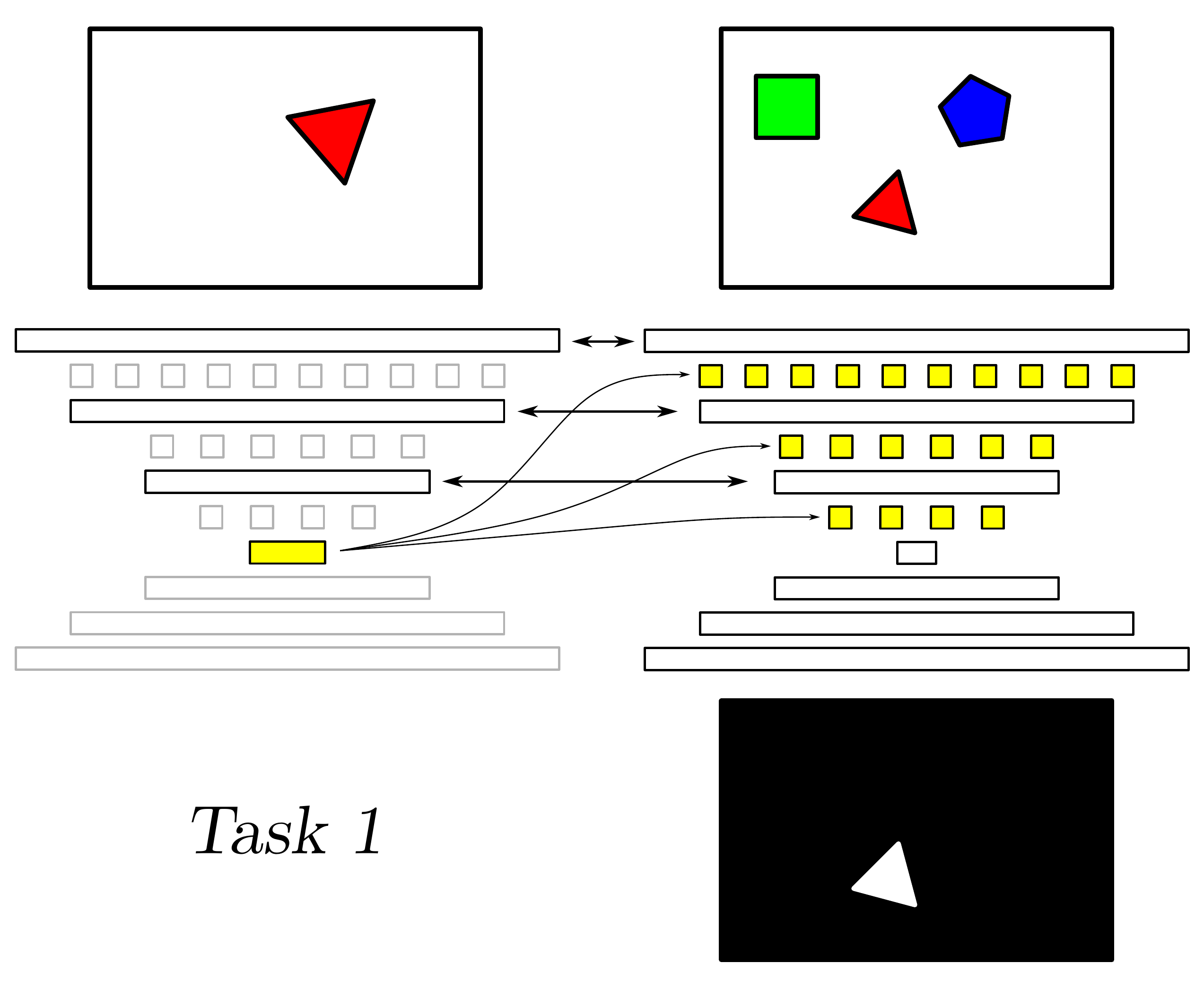}
	\caption{Our approach feeds the \emph{needle} image through the convolutional stage first, achieving a $4,544$ dimensional weight vector. This weight vector is used to weight each filter output during the convolutional deconstruction of the \emph{haystack} image. Both networks displayed share their convolutional parameters. Reconstruction of the \emph{needle} object on the \emph{haystack} image denotes the main task, \emph{Task 1}, of the network to be learned.}
	\label{fig:schematics2}
\end{figure}

First, the \emph{needle} image is presented to the feature extraction part of the network, i.e., the convolutional layers. A $4,544$ dimensional output vector is returned, providing an additional $4,544$ filter weights $w_i$ for extraction.
Next, the \emph{haystack} image is presented to the network. This time, every one of the $4,544$ convolutional feature extraction filters is weighted by the weight $w_i$ extracted from the \emph{needle} image. This effectively allows the network to block the output of any filters that detect objects other than the one looked for. As the convolutional filter parameters are shared, the amount of variables to be learned is still manageable for graphics cards commonly used in deep learning.
The dimensions of the \emph{haystack} image are then reconstructed using deconvolutional layers, and a belief map of equal size is generated, highlighting the areas for which the network extracted similar features to the \emph{needle} image.  Figure~\ref{fig:schematics2} shows a simplified version of the approach. The belief map is then compared to the desired segmentation, and an error value is returned. The IoU measured between generated belief map and ground truth forms the main loss to be minimized. Other error metrics are extracted as well, as detailed in the following section.

\subsection{Error Measures}
\begin{figure}[tb]
	\centering
	\includegraphics[width=1\linewidth]{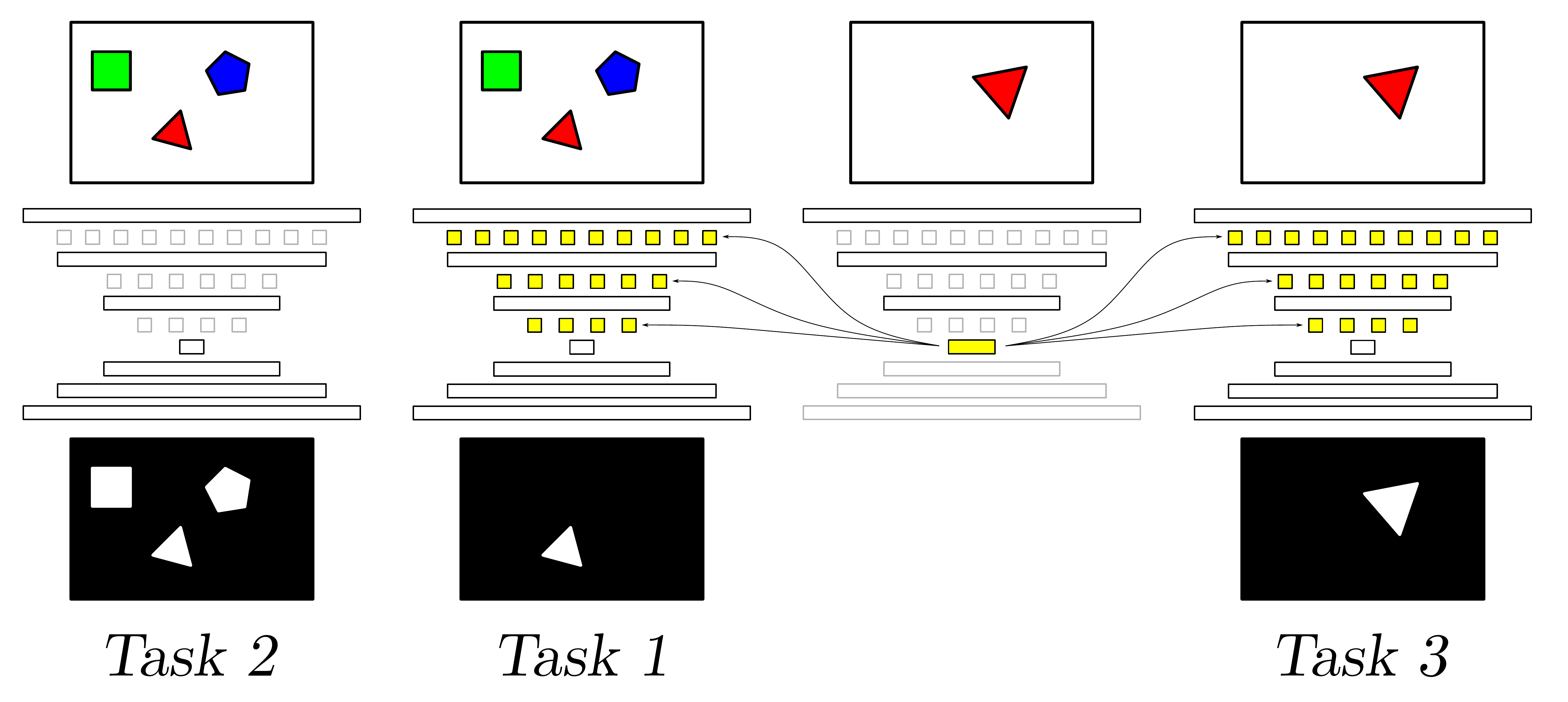}
	\caption{Multiple error measures can be derived from a single \emph{needle/haystack} image combination. Learning multiple tasks from multiple error measures can provide more information to the optimizer every iteration. First, the constrained network should detect both instances of the \emph{needle} object with the same weights. Second, the unconstrained network should detect every object, effectively learning a background model to be removed. The third task forces the network to extract features from the desired region only, i.e., from the \emph{needle} object.}
	\label{fig:schematics3}
\end{figure}

In order to aid the network to produce a highly abstract representation and to speed up convergence~\cite{caruana1998multitask}, multiple error measures are combined. This transforms the segmentation task into a multi-task problem, which have been shown to be able to generate higher levels of abstraction than single-task learning. Figure~\ref{fig:schematics3} illustrates how three different IoU metrics are obtained and subsequently minimized during each training step.
The first independent task for the network to learn is to extract the \emph{needle} object from the \emph{haystack} image, as described in Section~\ref{sec:netDesign} above.
The second task is trained at the same time, being that the network should extract all the objects of the \emph{haystack} image if no weights are provided, i.e., all filters are in effect.
This combination aids the network in learning narrow-band filters, that only extract the required object type and ignore the background fully.
Finally, the third task to be learned is making the network re-detect the \emph{needle} object with activated weights. This task forces the network to learn filters at least wide enough to extract both  \emph{needle} objects in both images, regardless of lighting or scale. Training all three tasks, i.e., combining the narrowing and widening effects, the network can learn filters that are only sensitive to a certain range of object variability, ignoring objects that are outside the learned range in feature space representation.

\subsection{Vector Spread Loss}
\label{sec:vector_spread}
In order to further aid the network learning a mapping of input images into a spanning feature space, we add an additional fourth task, minimizing a \emph{vector spread} loss term. The aim of the vector spread error metric is to minimize intra-class distances in feature space, while maximizing inter-class distances, thus spreading feature vectors of different classes. Inspired by the angular error used by~\cite{wang2017deep}, we model the vector spread error using a cosine distance, as shown in Equation~\ref{eq:cosine_distance}, between object representation vectors in feature space.
For every mini-batch of size $b$, the vector spread error is calculated for all the combinations of feature vectors gathered from the last convolutional layer.
Equation~\ref{eq:vector_spread} details the calculation of the vector spread error metric.
The distance of every feature vector combination of $X_i$ and $Y_j$ is desired to be $1$ if the classes of $X_i$ and $Y_j$ differ, or $0$ if $X_i$ and $Y_j$ belong to the same class.

\begin{equation}
\label{eq:cosine_distance}
CosineDistance(X, Y) = \frac{cos^{-1}(\frac{X \cdot Y}{\|X\| \|Y\|})}{\pi}
\end{equation}

\begin{multline}
\label{eq:vector_spread}
\mathcal{L}_{VectorSpread} = \sum\limits_{i=0}^{b}\sum\limits_{j=0}^{b}\frac{D_{ij}}{b\times b} \text{ , where} \\ D_{ij} = 
\begin{cases}
    CosineDistance(X_i, Y_j) & \text{if } Class(X_i) = Class(Y_j)\text{,} \\
    1 - CosineDistance(X_i, Y_j) & \text{otherwise.}
\end{cases}
\end{multline}
\vspace{0.2cm}

\begin{figure}[tb]
	\centering
	\includegraphics[width=1.0\linewidth]{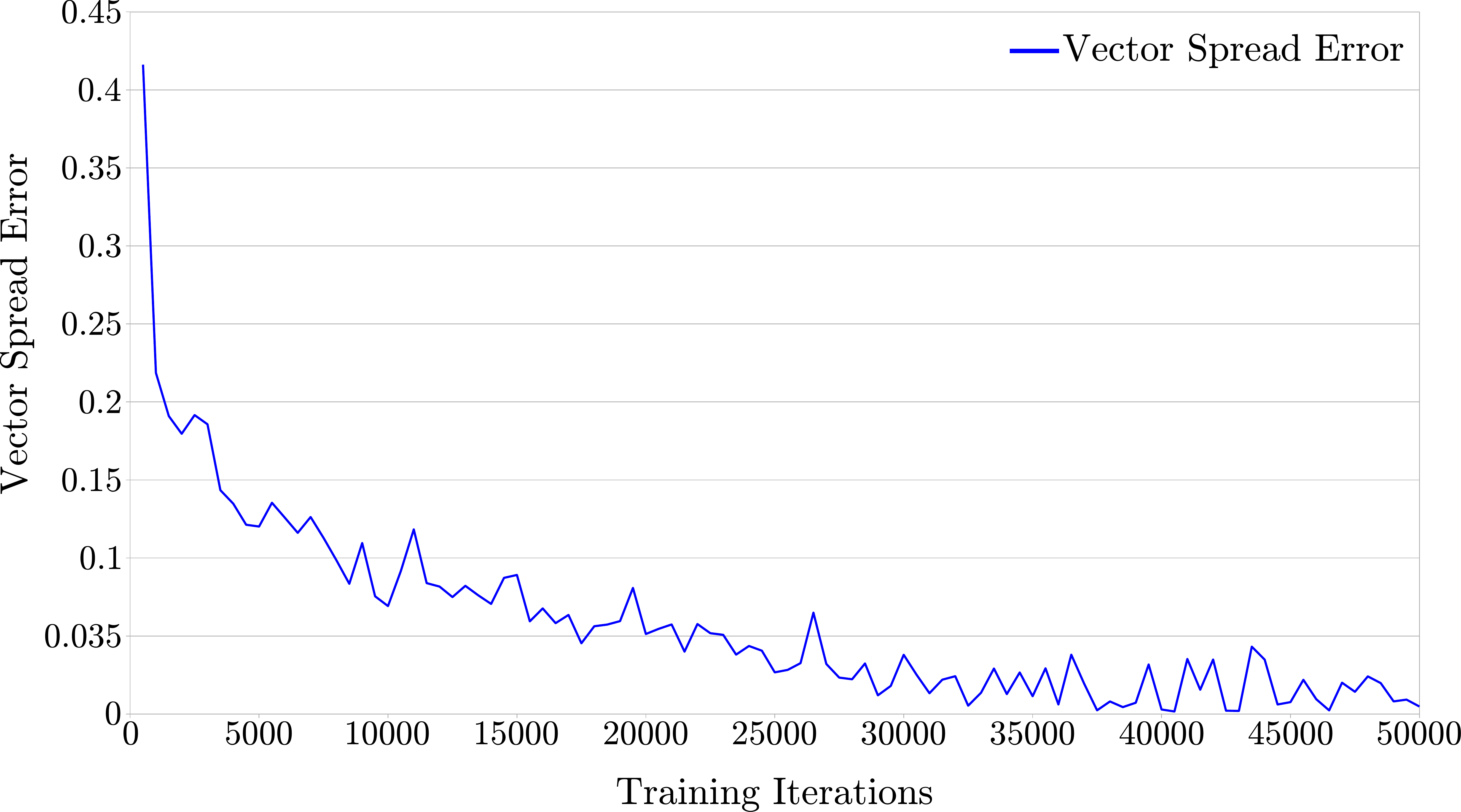}
	\caption{The decreasing vector spread error during training.}
	\label{fig:vectorSpreadChart}
\end{figure}

The differences between calculated and desired values are summed up and normalized by dividing by the number of elements, $b\times b$, forming the vector spread loss term. This metric favors a distinct set of filters being activated for every class encountered by forcing their vector representations to be orthogonal in feature space. As filters shared for multiple object types will extract both objects at all times, learning separate filters for different object types is an advantage. Additionally, it forces the network to collapse feature space representations of objects of the same class. Figure~\ref{fig:vectorSpreadChart} presents an example of the vector spread metric converging to a value near zero while training, indicating that the dimensionality of the feature space was high enough to fit all the classes collision free.

\subsection{Weighting Loss Terms}
\begin{figure}[tb]
	\centering
	\includegraphics[width=0.7\linewidth]{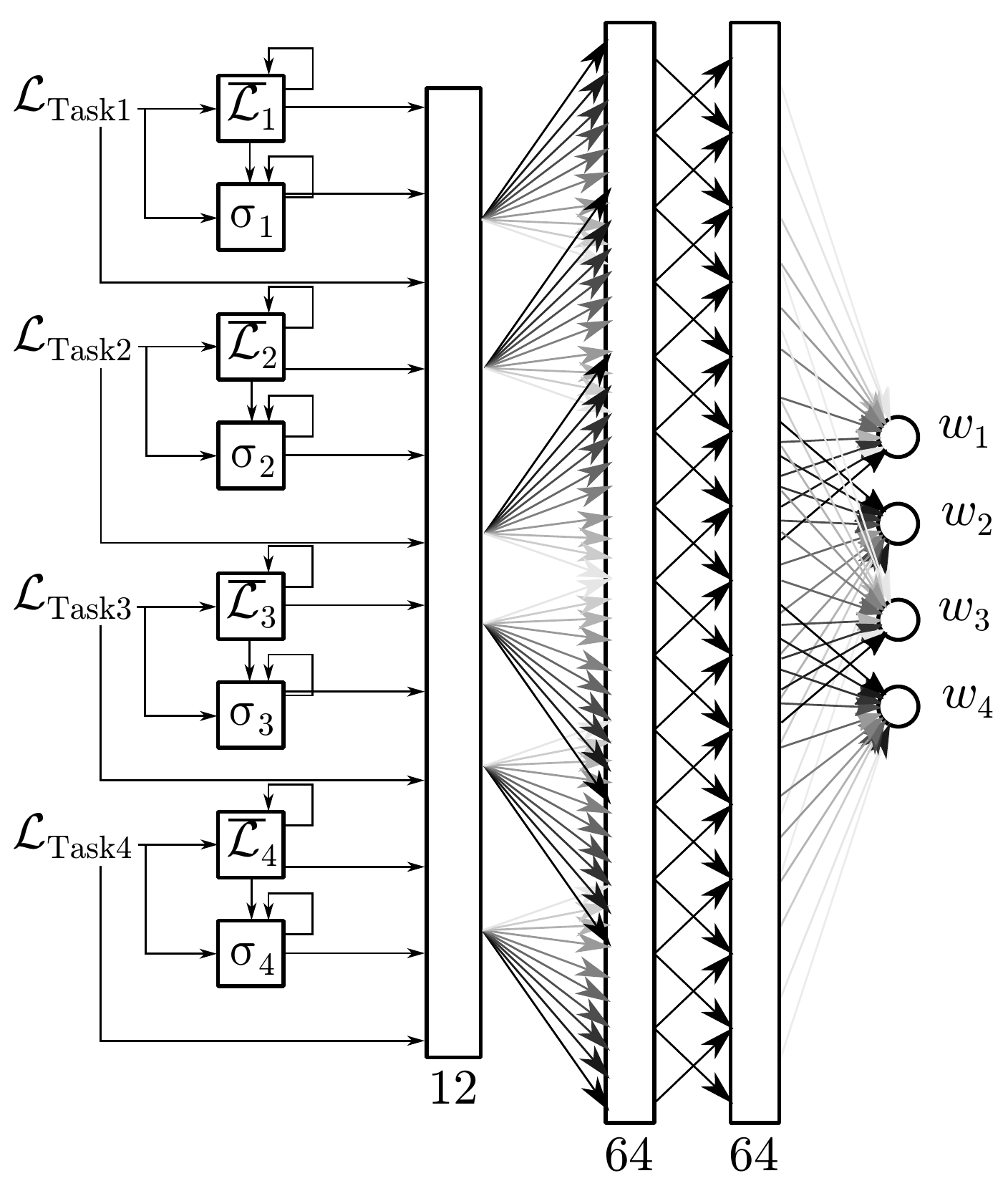}
	\caption{The layout of the auxiliary network to produce weights for every loss function used. The loss values for the four different tasks are averaged over time, and their variances are measured. The current loss values, average values and their variances are fed into the network, after which two fully connected layers produce 4 output weights.}
	\label{fig:auxnet}
\end{figure}

Inspired by and extending the work of \cite{schnieders2018fast}, we implemented an auxiliary neural network to produce an optimal weighting between the different loss terms. A network taking the current losses, their exponential moving averages and variances as inputs and featuring two fully connected layers of \numprint{64} nodes each produced quickest convergence based on limited trials. Figure~\ref{fig:auxnet} shows the layout of the auxiliary network. All activation functions are leaky ReLUs, and the same optimizer settings are used as for the main DCNN. Training the auxiliary network is done as detailed in~\cite{schnieders2018fast}.

\section{Experiments}
\label{sec:exp}
We train a deep convolutional neural network on the new, publicly available {\rc} dataset. We split up the dataset into \numprint{11} object classes to be used for training, and \numprint{2} object classes to be held out during training. The network is trained on all the \numprint{78} combinations of different classes to be held out. For every combination, we train the network for \numprint{50000} iterations, as initial tests have shown the network converging well within this span. The evaluation set contains \emph{needle} images featuring exclusively held out classes, and \emph{haystack} images featuring at least both held out object classes, and up to \numprint{10} known objects of known classes as distractors.


\emph{He} initialization, as proposed in~\cite{he2015delving}, is used to initialize all weights of the neural network. Due to memory constraints of the GPUs used, no larger mini-batch size than \numprint{6} could be employed. The established Adam optimizer~\cite{kingma2014adam} is used to minimize the loss term, with a learning rate of $\num{1e-5}$, $\beta_1=0.9$, $\beta_2=0.999$, and $\epsilon=\num{1e-08}$.
Every \numprint{500} iterations during training, we evaluate the network's capability to extract unknown classes on the entire evaluation set. This allows us to produce charts detailing the learning progress.
In order to evaluate the performance gains in abstractive power and convergence speed, we repeat the experiments with the vector spread error disabled, and on a singe IoU error for the first identified task only.
Results of these tests can be found in Section~\ref{sec:results}.

\subsection{Datasets}
\label{sec:datasets}
While a multitude of object segmentation datasets exists~\cite{garciaGagcia2017review,janai2017computer}, almost none include industrial objects, an exception being T-LESS~\cite{hodan2017tless}. T-LESS includes \numprint{38000} training and \numprint{10000} evaluation RGBD images of \numprint{30} industrial object types.
However, due to the nature of the dataset, all objects are texture-less and of the same color.

\begin{figure}[tb]
	\centering
	\includegraphics[width=\linewidth]{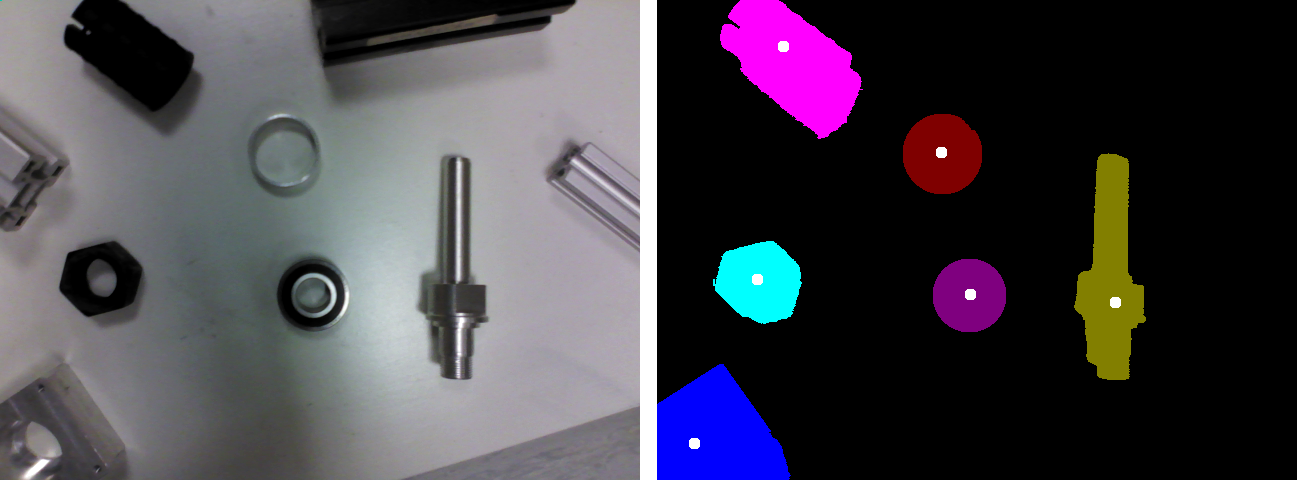}
	\caption{Left: An example image from the {\rc} dataset. Right: Automatically acquired segmentation and manually labeled pickup points. Note that the automatic segmentation frequently includes shadows with the object.}
	\label{fig:rcExample}
\end{figure}
\begin{figure}[tb]
	\centering
	\includegraphics[width=0.32\linewidth]{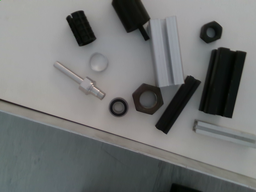} \includegraphics[width=0.32\linewidth]{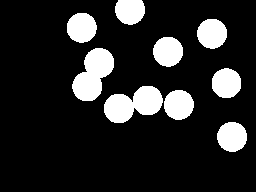} \includegraphics[width=0.32\linewidth]{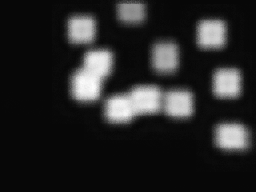}
	\caption{Left: An example \emph{haystack} image from the evaluation set. Center: The ground truth, approximated segmentation for learning Task~2, i.e., all filters are active. Right: The segmentation output of the network.}
	\label{fig:manyobjs}
\end{figure}

The {\rc} dataset~\cite{schnieders2018fast} currently contains \numprint{36000} RGBD images of industrial objects, of which \numprint{15800} are manually labeled with a center of gravity, or \emph{pickup point}. The objects featured are the \numprint{13} object classes used in the annual {\rc} challenge~\cite{kraetzschmar2014robocup}. As object detection and picking performance on this dataset can be tested with our robotic setup, most of our results are derived from this dataset. Figure~\ref{fig:rcExample} shows an example of this dataset and segmentation derived from previously world-cup winning {\rc} software~\cite{broecker2014winning}. As the automatic segmentation fails when objects are placed closer than approximately \SI{3}{\cm} apart, we obtained an alternative, approximate segmentation by extracting a circular region around the pickup point instead. This way, every object is represented by the same amount of pixels, preventing the tendency to select larger objects frequently happening when training on IoU~\cite{shaban2017one}. Figure~\ref{fig:manyobjs} shows an example of the approximated segmentation labels used, and the output of the trained network.


\begin{figure}[tb]
	\centering
	\includegraphics[width=\linewidth]{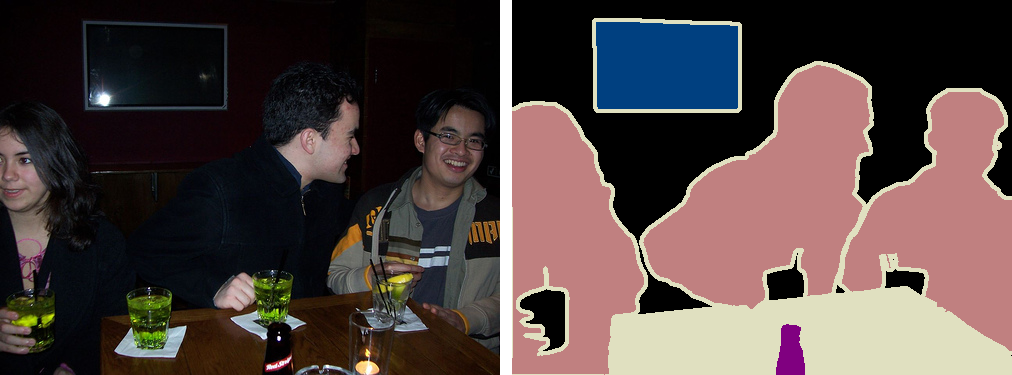}
	\caption{Left: An example image taken from the Pascal VOC dataset featuring three instances of the person object, one screen, and one bottle. Right: The corresponding ground-truth segmentation.}
	\label{fig:pascalVOCExample}
\end{figure}

\begin{figure}[tb]
	\centering
	\includegraphics[width=1.0\linewidth]{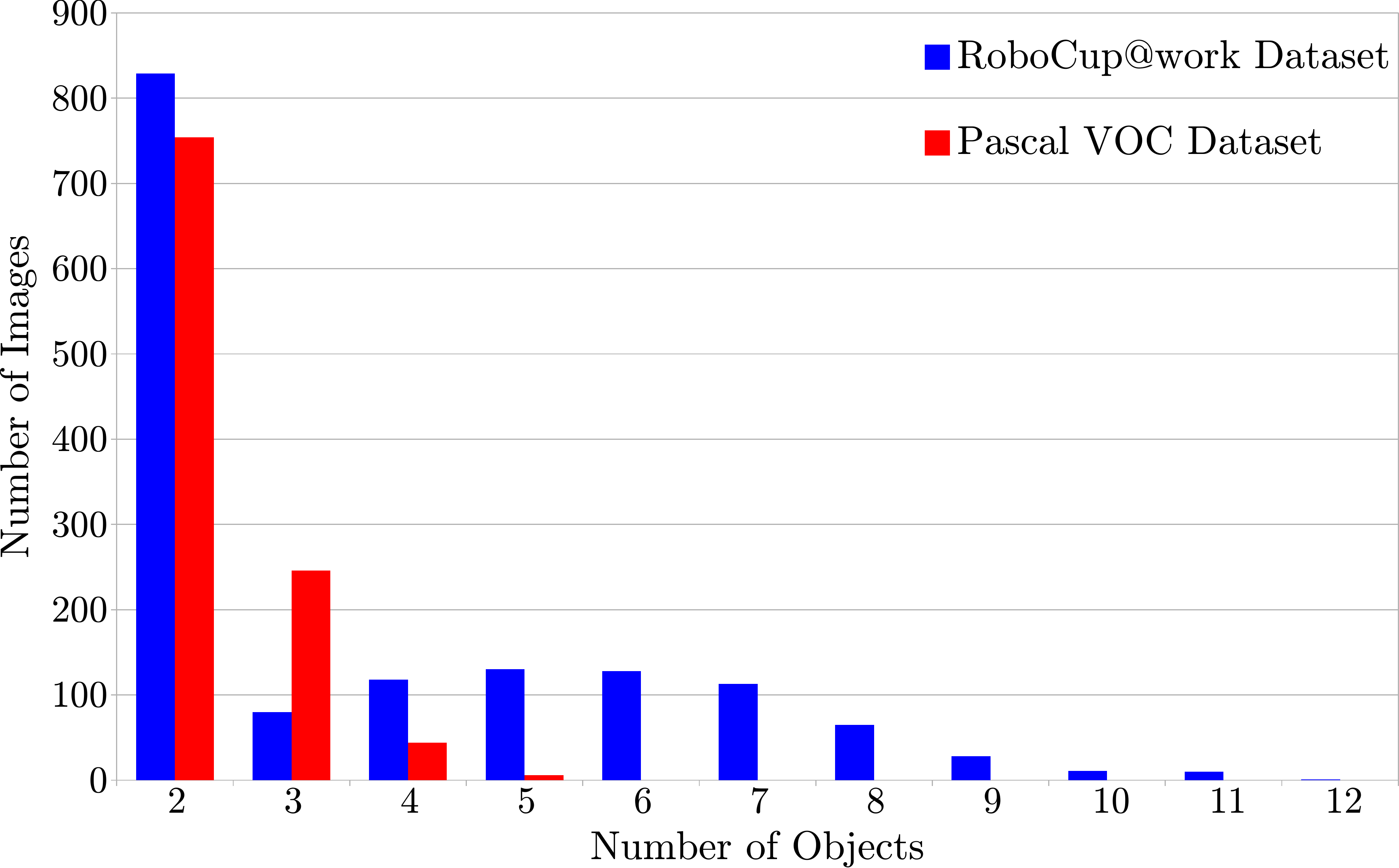}
	\caption{A histogram of object counts in the {\rc} (blue) and Pascal VOC (red) datasets, with the X axis enumerating the number of occurring objects, and the Y axis denoting the number of images featuring this number of objects. Single object images excluded for clarity.}
	\label{fig:dataset_compare}
\end{figure}

At the time of writing, the most widely used image segmentation dataset is the Pascal VOC dataset~\cite{everingham2010pascal}, featuring over \numprint{2900} RGB images semantically segmented into \numprint{20} object classes. The object classes range from outdoor vehicles, over animals, to objects commonly found indoors. Depth information is not available, and the intra-class variation is high, as is to be expected for real world everyday objects. This makes the Pascal VOC dataset very challenging. Figure~\ref{fig:pascalVOCExample} shows an example image featuring multiple classes and its corresponding ground-truth segmentation.
Using the Pascal VOC dataset as an independent benchmark for one-shot segmentation poses yet another challenge. To ensure the network learns a true one-shot object segmentation rather than a zero-shot foreground detection based on saliency~\cite{hong2015online}, the \emph{haystack image} has to include multiple objects, so that the network can be penalized for extracting both. The Pascal VOC dataset only contains \numprint{1050} possible \emph{haystack} images, with the majority of those images containing just two object classes. While the same is true for the {\rc} dataset, not only does it contain more (\numprint{1513}) possible \emph{haystack} images, but the distribution of object classes is much more favorable, as shown in Figure~\ref{fig:dataset_compare}. The more different objects are present in a \emph{haystack}-image, the more information is available per iteration for the network to learn. Additionally, the detection task becomes harder, as it is more likely to mislabel an object. \emph{Haystack} images containing multiple objects can be used in conjunction with every class they feature, further increasing the number of possible \emph{needle/haystack} combinations to train.

\section{Results}
\label{sec:results}
\begin{figure}[tb]
	\centering
	\includegraphics[width=\linewidth]{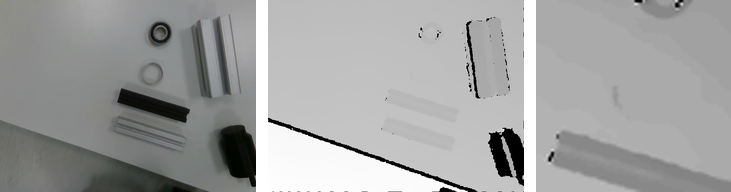}
	\caption{The \emph{Distance Tube} object (small grey aluminium ring) next to other objects in visible light (left) and as occurring on the depth channel (center). The right image shows a contrast enhanced magnification of the small portion of the distance tube that is visible on the depth channel.}
	\label{fig:distance_tube_rgbd_problem}
\end{figure}

\begin{figure}[tb]
	\centering
	\includegraphics[width=0.49\linewidth]{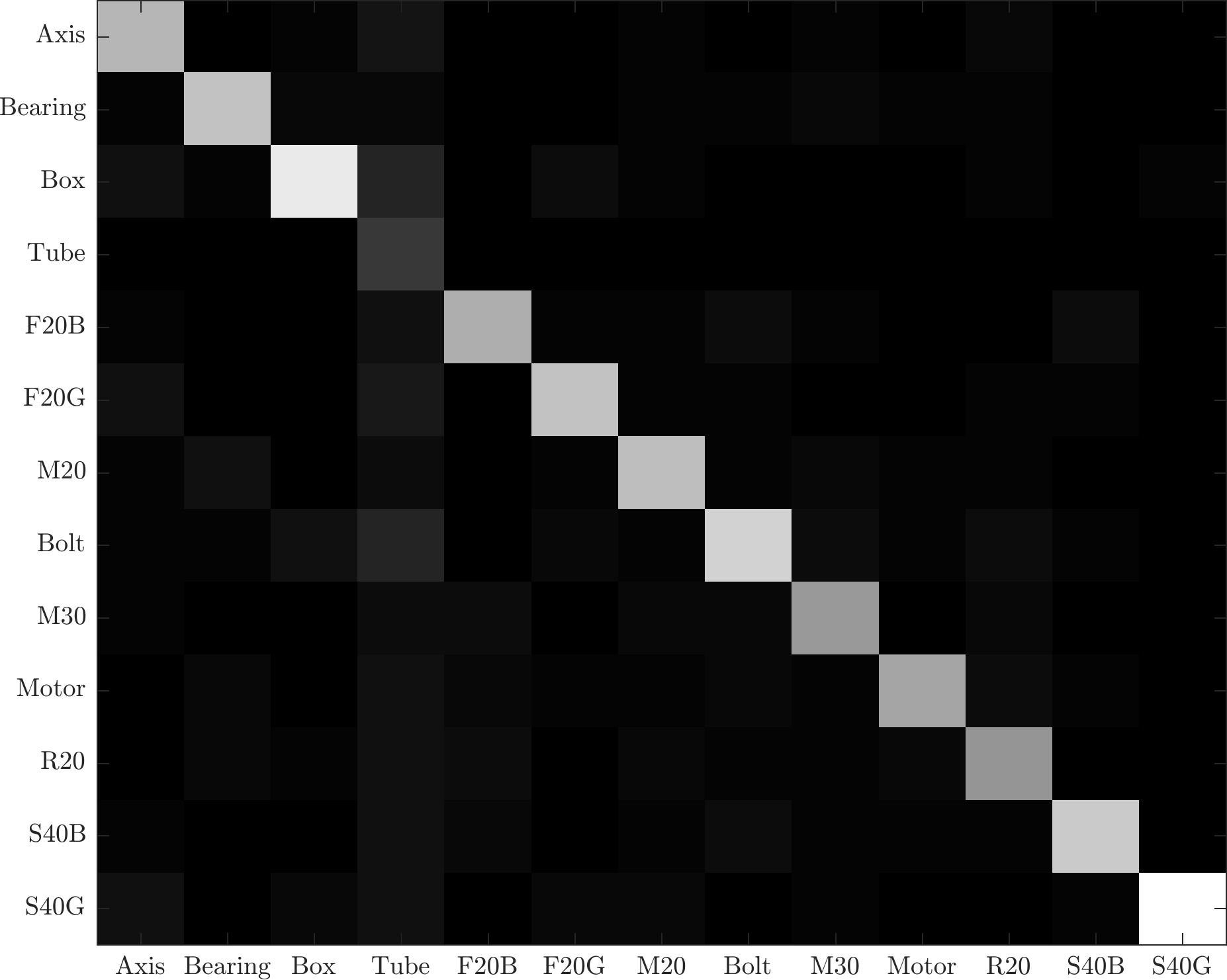}\hfill\includegraphics[width=0.49\linewidth]{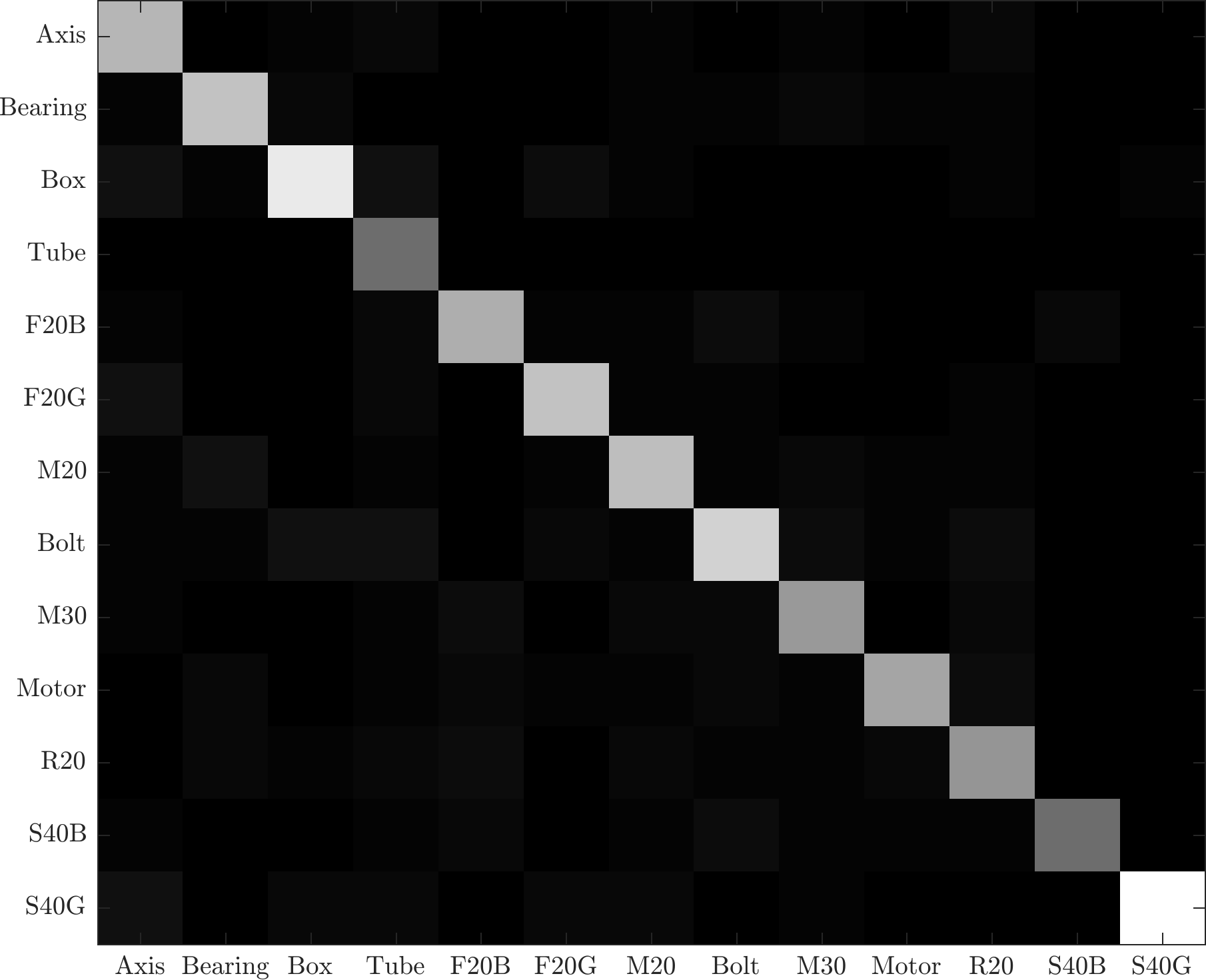}
	\caption{Left: The confusion matrix of correctly one-shot detecting the \emph{needle} object in the \emph{haystack} image when training on all RGBD channels. The \emph{Distance Tube} object shows a very low recall of {18.5\percent}. As it is the only object not showing up in the depth channel, training the network without it results in overfitting on depth features. Right: The same confusion matrix, when trained on RGB information only.}
	\label{fig:confusion_rgbd}
\end{figure}

We obtain classification data by analyzing if combining the prediction with the desired segmentation mask produces a higher IoU measure than combining it with any other object segmentation mask.
Training the network on all \numprint{78} possible combinations of included and held out objects in the {\rc} dataset resulted in an averaged $69\percent$ classification rate of previously unseen data. Investigating the confusion matrix, shown in Figure~\ref{fig:confusion_rgbd} (left), quickly identifies recalling the \emph{Distance Tube} object as the weakest contributor. The low recall of $18.5\percent$ on this object shows the main limitation of this and any likewise operating one-shot object detectors, as the object inhabits a point outside the learned feature space, spanned by training on the remaining objects. This is due to a unique feature of the \emph{Distance Tube}, being near invisible on the depth channel, as shown by Figure~\ref{fig:distance_tube_rgbd_problem}. This object being held out, the network overtrains on the depth channel otherwise being a good indicator for object presence, and resorts to marking another object in the \emph{haystack} image. Removing the \emph{Distance Tube} object from the dataset raises the accuracy to $81.15\percent$.

\begin{figure}[tb]
	\centering
	\includegraphics[width=1\linewidth]{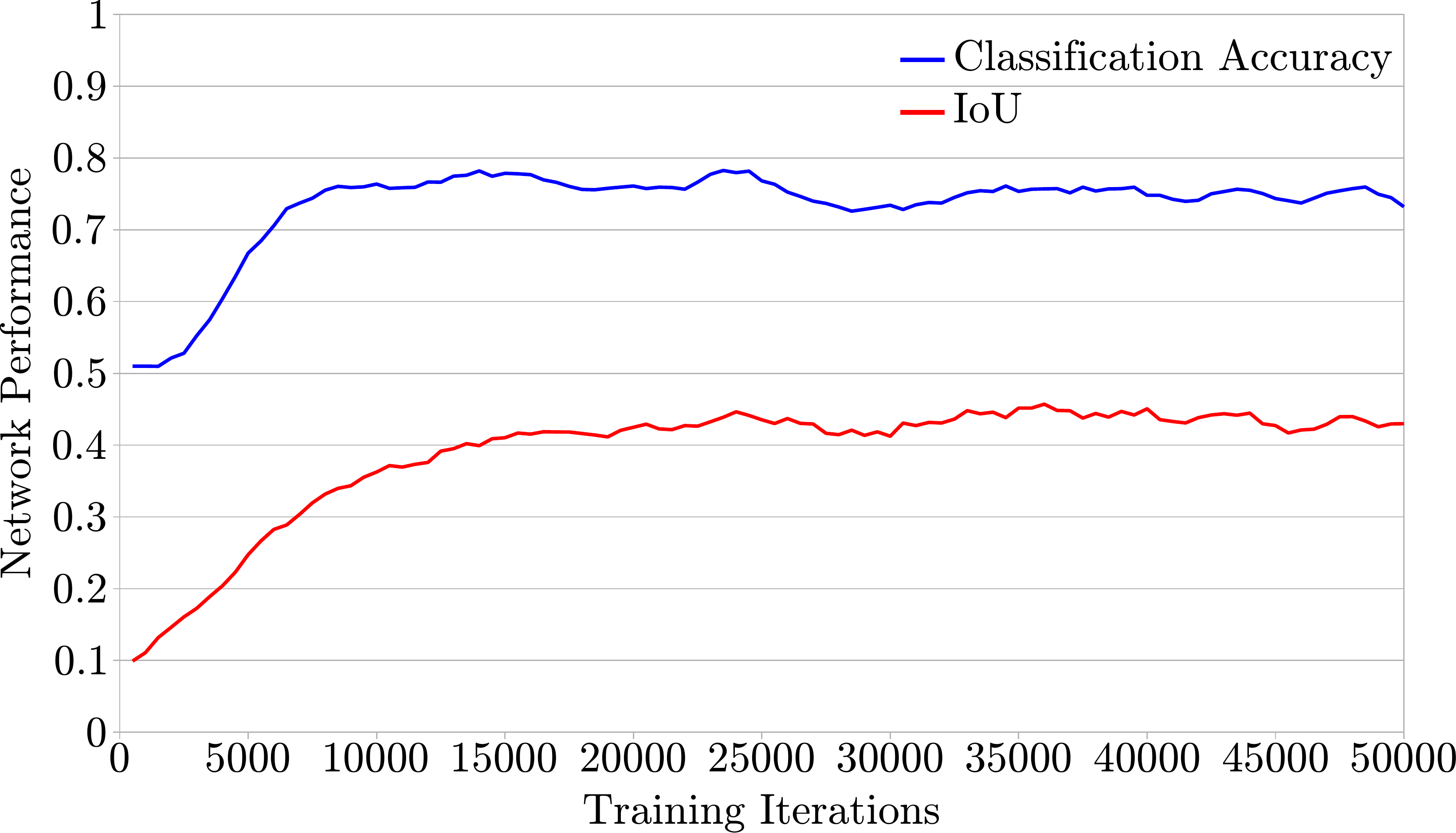}
	\caption{Classification accuracy (blue) and IoU (red) obtained on untrained classes averaged over the 78 training sessions and plotted for every 500 training iterations. Training and validation data contained RGB images only.}
	\label{fig:acc_iou_chart}
\end{figure}

Disabling the depth channel forces the network to take only RGB information into account, and raises the recall of the \emph{Distance Tube} to $48.9\percent$. This configuration leads to an overall accuracy of $73.1\percent$, and an average IoU of $0.45$ after convergence. Using the pickup error metric presented in~\cite{schnieders2018fast}, we estimate the probability for successfully picking an identified object. The one-shot pickup rate achieved is $87.53\percent$. Figure~\ref{fig:acc_iou_chart} shows the learning progress of the trained classifiers by plotting the accuracy and IoU metrics obtained on the held out validation set every \numprint{500} steps during training.
Figure~\ref{fig:confusion_rgbd} (right) shows the according confusion matrix, and Table~\ref{tab:precision_recall_rgb} lists the precision and recall measures per object type.

\begin{table}[t]
\centering
\caption{Precision and Recalls trained on RGB}
\label{tab:precision_recall_rgb}
\begin{tabular}{lcc}
\toprule
Object class & Precision & Recall \\
\midrule
Axis & 0.78 & 0.65 \\
Bearing & 0.78 & 0.75 \\
Bearing~Box & 0.74 & 0.80 \\
Distance~Tube & 0.92 & 0.49 \\
F20\_20\_B & 0.73 & 0.75 \\
F20\_20\_G & 0.75 & 0.78 \\
M20 & 0.73 & 0.70 \\
M20\_100 & 0.68 & 0.75 \\
M30 & 0.70 & 0.68 \\
Motor & 0.69 & 0.79 \\
R20 & 0.66 & 0.65 \\
S40\_40\_B & 0.61 & 0.78 \\
S40\_40\_G & 0.78 & 0.91 \\
\bottomrule
\end{tabular}
\end{table}



\begin{figure}[tb]
	\centering
	\includegraphics[width=1\linewidth]{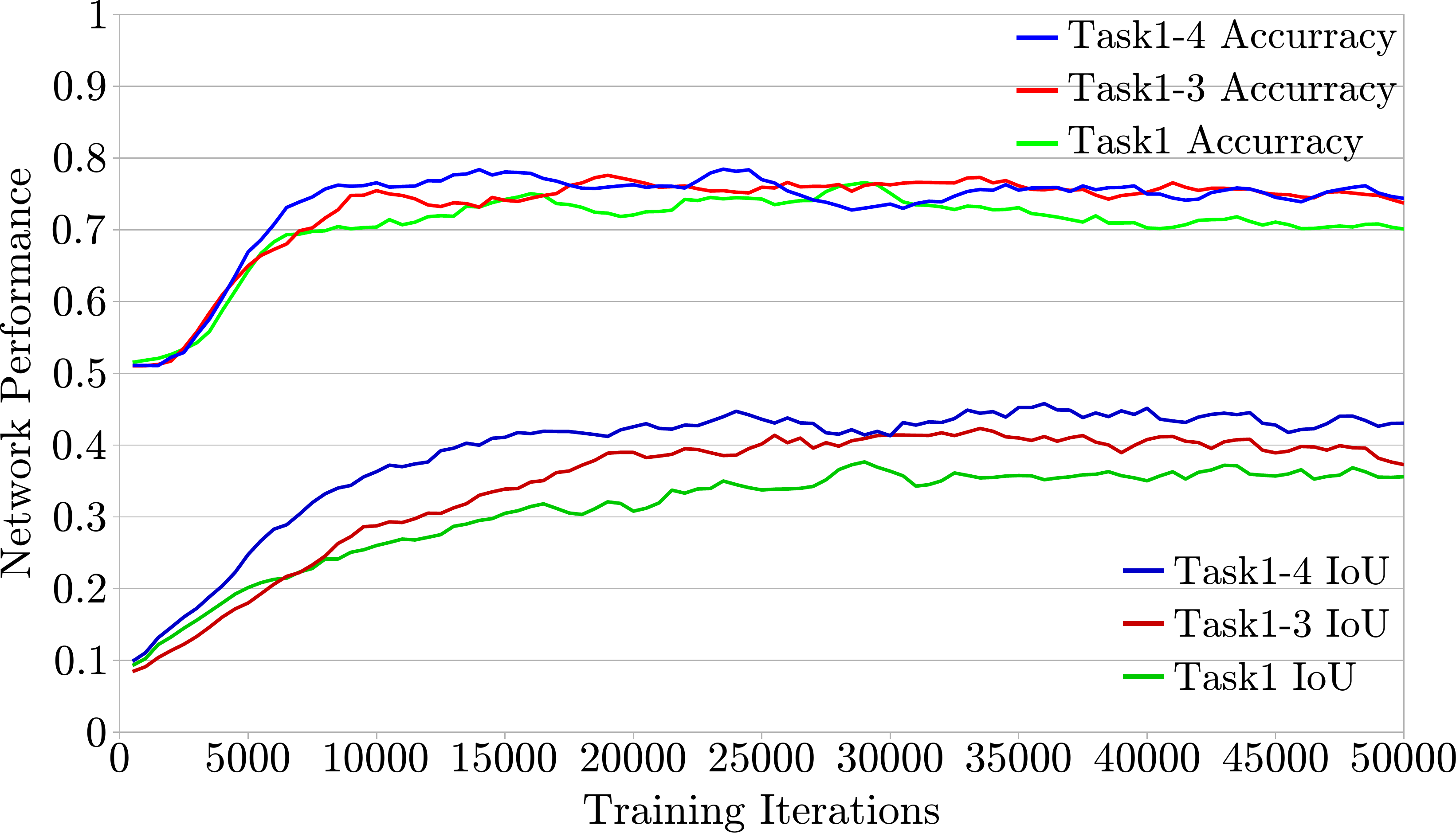}
	\caption{Accuracy and IoU on the evaluation set during training for three training configurations. Blue: All 4 Tasks identified are trained. Red: Tasks 1-3 were trained. Green: Only IoU was trained. The top three charts indicate accuracies, the lower charts IoU.}
	\label{fig:acc_iou_3methods}
\end{figure}

We repeated the tests with disabled Task 4, and with disabled Tasks 1-3. The resulting IoU and accuracy curves are shown in Figure~\ref{fig:acc_iou_3methods}. Training on all 4 Tasks converges faster and to a lower error than training on fewer. Table~\ref{tab:multitask_results} provides the convergence points and average accuracy after convergence. We observe that adding the \emph{vector spread} metric does not influence the reached accuracy in a significant manner, but, compared to training only on Tasks 1-3, reduces the iterations to convergence by $22.22\percent$.

\begin{table}[t]
\centering
\caption{Multi-task learning effect}
\label{tab:multitask_results}
\begin{tabular}{lcc}
\toprule
Tasks Trained & Iterations to converge & Avg. Accuracy \\
\midrule
Only IoU & \numprint{10500} & 72.40\percent \\
Tasks 1-3 & \numprint{9000} & 75.31\percent \\
Tasks 1-4 & \numprint{7000} & 75.47\percent \\
\bottomrule
\end{tabular}
\end{table}

\begin{figure}[tb]
	\centering
	\includegraphics[width=1\linewidth]{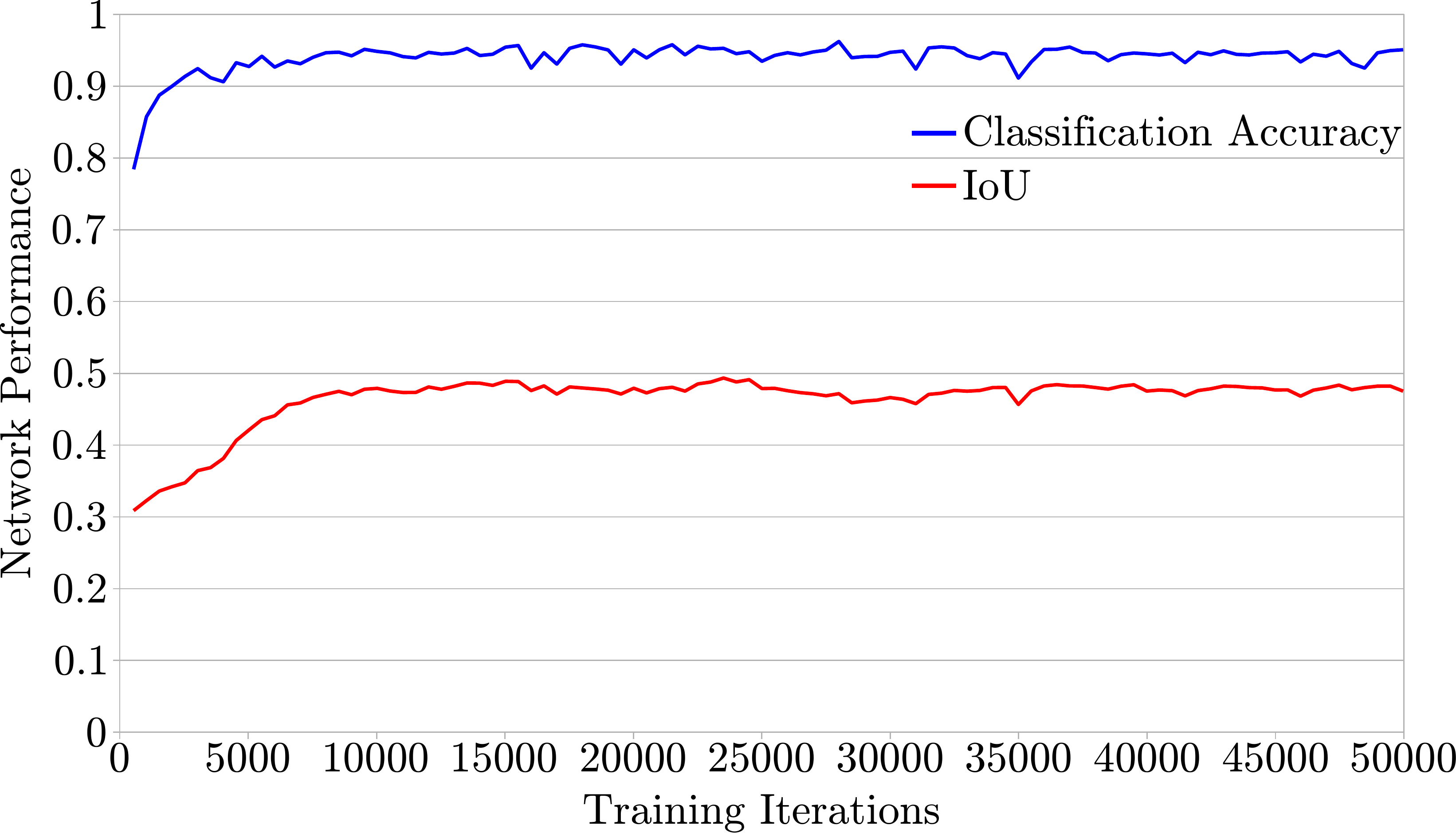}
	\caption{Classification accuracy (blue) and IoU (red) obtained on known classes, plotted for every 500 training iterations. Training and validation data contained RGB images only.}
	\label{fig:acc_iou_chart_on_all}
\end{figure}

To evaluate the performance of the network when re-trained with all the available information, we repeated the network training with no classes withheld and instead splitting the dataset into a training and evaluation set at the ratio of $4$ to $1$. Figure~\ref{fig:acc_iou_chart_on_all} shows the rapid learning of the network, achieving peak performance of $94\percent$ correct classification after as few as \numprint{4000} iterations. The pickup rate for correctly identified objects is estimated to be $97.77\percent$. This clearly shows the advantage of re-training over one-shot segmentation, and further confirms the necessity to be able to re-train as quickly as possible.

\section{Conclusion}
\label{sec:conc}

In this paper, we proposed a novel framework for one-shot object segmentation using fully convolutional Siamese Deep Neural Networks. We demonstrated its effectiveness in segmenting new objects in industrial settings, and its rapid re-training capabilities, to replace one-shot segmentation with more accurate fully trained segmentation as quickly as possible.
We have shown that our approach can successfully extrapolate how to segment unknown object types based on training and evaluating it on different object classes.
We have also shown the limitations of this approach, being that it can overfit on specific details prevalent in the training set. 
However, this limitation can easily be overcome in an industrial setting, where a true spanning set of object classes can be combined to form a training set, allowing the network to robustly detect and segment previously unknown objects based on a single example image.

By applying multi-task learning techniques, our network can converge after as few as \numprint{4000} training iterations. In an industrial setting, our approach can be used as a one-shot segmentation network to pick up objects while re-training is in effect. After the very short training time of \numprint{57} minutes, a $94\percent$ accurate classifier can pick up identified objects with an estimated rate of $97.77\percent$. Before re-training is completed, the network can robustly operate in a less accurate, one-shot segmentation manner, fully alleviating any training time bottleneck.
We therefore conclude that robots in the \emph{Factory of the Future} can greatly benefit from our presented one-shot object segmentation approach, allowing them to autonomously produce, locate, and pick novel objects without human supervision.

Our future work will extend the one-shot segmentation to a few-shot segmentation approach, in which multiple shots of the same prototype object are provided. Other research~\cite{shaban2017one} has shown that object identification can greatly benefit from this transition. Further future work includes extending the {\rc} dataset with arbitrary backgrounds and decoy objects, and providing full segmentation data. As this will provide an even better \emph{Factory of the Future} approximation, our approach will be re-evaluated on the extended dataset and comparable datasets such as T-LESS~\cite{hodan2017tless}. The findings in the paper will also be applied in pick-and-place tasks on real robot platforms, possibly with interactive perception \cite{bohg2017interactive} and other sensing inputs (e.g., tactile sensing \cite{luo2017robotic}).

\section{Acknowledgement}

We are very grateful to James Butterworth and Jacopo Castellini for helpful comments and discussions. This work was supported by the EPSRC project ``Robotics and Artificial Intelligence for Nuclear (RAIN)" (EP/R026084/1).